\documentclass[journal]{IEEEtran}

\ifCLASSINFOpdf
\else
   \usepackage[dvips]{graphicx}
\fi
\usepackage{url}
\usepackage{amsmath} % for \boldsymbol macro 
\usepackage{bm} % for \bm macro
\usepackage{amsfonts}
\usepackage{amssymb}
\usepackage{balance}
\usepackage{cite}
\usepackage{flushend}
\usepackage{subfigure}
\usepackage{booktabs}
\usepackage{tabularx}
\usepackage{multirow}

\usepackage{graphicx}

\begin{document}

\title{GA-NET: Global Attention Network for Point Cloud Semantic Segmentation}

\author{Shuang Deng and Qiulei Dong
\thanks{This work was supported by the National Natural Science Foundation of China (Grant Nos. U1805264 and 61991423), the Strategic Priority Research Program of the Chinese Academy of Sciences (XDB32050100), and the Open Research Fund from Key Laboratory of Intelligent Infrared Perception, Chinese Academy of Sciences.}
%\thanks{S. Deng is with the School of Artificial Intelligence, University of Chinese Academy of Sciences, Beijing 100049, China, and also with the National Laboratory of Pattern Recognition, Institute of Automation, Chinese Academy of Sciences, Beijing 100190, China (e-mail: shuang.deng@nlpr.ia.ac.cn).}
\thanks{S. Deng and Q. Dong are with the National Laboratory of Pattern Recognition, Institute of Automation, Chinese Academy of Sciences, Beijing 100190, China, and also with the School of Artificial Intelligence, University of Chinese Academy of Sciences, Beijing 100049, China, and also with the Center for Excellence in Brain Science and Intelligence Technology, Chinese Academy of Sciences, Beijing 100190, China (e-mail: shuang.deng@nlpr.ia.ac.cn; qldong@nlpr.ia.ac.cn).}
}

\markboth{Submitted to IEEE Signal Processing Letters}
{Shell \MakeLowercase{\textit{et al.}}: Bare Demo of IEEEtran.cls for IEEE Journals}
\maketitle

\begin{abstract}
How to learn long-range dependencies from 3D point clouds is a challenging problem in 3D point cloud analysis. Addressing this problem, we propose a global attention network for point cloud semantic segmentation, named as GA-Net, consisting of a point-independent global attention module and a point-dependent global attention module for obtaining contextual information of 3D point clouds in this paper. The point-independent global attention module simply shares a global attention map for all 3D points. In the point-dependent global attention module, for each point, a novel random cross attention block using only two randomly sampled subsets is exploited to learn the contextual information of all the points. Additionally, we design a novel point-adaptive aggregation block to replace linear skip connection for aggregating more discriminate features. Extensive experimental results on three 3D public datasets demonstrate that our method outperforms state-of-the-art methods in most cases.
\end{abstract}


\IEEEpeerreviewmaketitle

\begin{figure*}[t]
	\centering
	\includegraphics[width=1.95 \columnwidth]{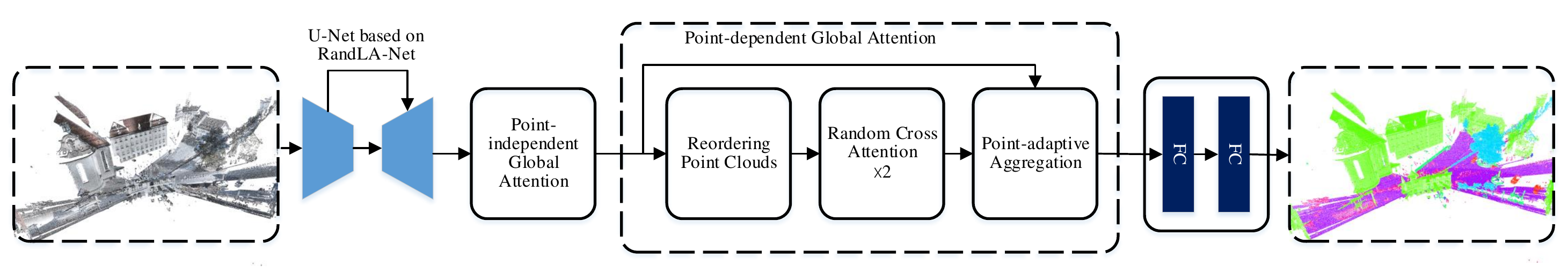}
	\caption{Architecture of the proposed GA-Net.}
	\label{fig1}
\end{figure*}

\section{Introduction}

%With 3D point cloud data being easily accessible in recent years, the techniques of 3D point cloud semantic segmentation have received increasing attention. Due to the proporty of unorderness and irregularity, how to exploit context information of point clouds using Deep Neural Networks(DNNs) is challenging.

\IEEEPARstart{T}{hree-dimensional} point cloud semantic segmentation is an important topic in the field of computer vision.
In recent years, a large amount of  Deep Neural Networks(DNNs)~\cite{2017pointnet, 2017pointnet++, 2018spidercnn, 2018dgcnn, 2018pointcnn, 2018pointsift, 2019pointweb, 2019gacnet, 2020randlanet, 2021scfnet, 2021rtn} for point cloud semantic segmentation have been proposed. Although these methods can capture the geometric structures of local regions well, the relationships between long-range neighborhoods of 3D point clouds are usually ignored.

In fact, the contextual information from long-range neighboring points is essential for 3D point cloud semantic segmentation. Some recent works~\cite{2020point2node, 2020mprm, 2020pointgcr} showed that the non-local module in NLNet~\cite{2018nlnet} could improve the performances of DNNs on point cloud segmentation significantly. 
%However, the entire point clouds act in their favours to be divided into blocks, and the non-local module is used on each small block. When the non-local module is used on entire point clouds of large 3D scenes, the computational complexity will increase enormously, which is impractical in some cases. 
However, due to the computationally expensive nature of the non-local module, these existing methods could not directly handle the input complete 3D scenes, but they have to split each input complete 3D scene into many small cubes in advance, and then use the non-local module to handle each small cube, resulting in an incomplete contextual feature. 
Although the non-local module is computationally complex, there are some methods~\cite{2019ccnet, 2020lrnnet} to simplify the non-local module in the field of image processing. However, since 3D point cloud is an unordered and irregular structure, these attention mechanisms cannot be applied to 3D point clouds directly.

In addition, the non-local module~\cite{2018nlnet} is point-dependent, where the calculated attention maps are dependent on different points. GCNet~\cite{2019gcnet} indicates that the point-independent attention method could also improve the capabilities of DNNs, where only one attention map is shared by all points. However, this method ignores the differences between local regions.

Addressing these problems, we propose an end-to-end global-attention-based model, named as Global Attention Network (GA-Net) for point cloud semantic segmentation with a moderate computational complexity. GA-Net consists a U-Net-based feature extractor and two modules called the point-independent global attention module and the point-dependent global attention module. 
In the point-independent global attention module, we compute an attention map and apply it to all points for obtaining point-independent global information. In the point-dependent global attention module, an efficient random cross attention block is designed to replace the non-local module~\cite{2018nlnet}, which is of lower computational complexity and could directly handle complete 3D scenes. Specifically, each point only has connections with two randomly sampled subsets, the number of which is much smaller than the entire point cloud. Besides, we design a novel point-adaptive aggregation block to replace linear skip connection for aggregating more discriminate features.
The point-independent global attention module aims to learn an attention map for extracting global but relatively coarse contextual features. Then, based on the features outputted from the point-independent global attention module, the point-dependent global attention module aims to extract more deliberate global contextual features for each 3D point respectively. 

%In the field of image processing, there are methods to simplify non-local module for computational efficiency. In CCNet~\cite{2019ccnet}, each position only has connections with horizontal and vertical directions in the image. Since 3D point cloud is an unordered and irregular structure, the attention mechanism in CCNet can not be applied to point clouds directly. GCNet~\cite{2019gcnet} creates a simplified network based on a query-independent attention formulation, in which only one attention map is shared with all points. This method obtains global information well, but ignores the differences between local regions. In this paper, we propose two modules named query-independent global attention module and the query-dependent ones for obtaining contextual information of 3D point clouds efficiently. Unlike GCNet only considering the query-independent contextual dependencies, we consider both of them. The first module computes a global attention map and shares it for all points. In the second module, we design a computational efficient attention mechanism named random cross attention block. Specifically, for each point, the contextual information of all the points is obtained through two randomly sampled subsets, the number of which is much smaller than the entire point cloud. Then, a point-adaptive aggregation block is used for aggregating more discriminate features. The lightweight property allows us to apply the two modules in a backbone network to construct our Efficient Global Attention Network (GA-Net). 

In sum, the main contributions of this paper include: 
\begin{itemize}
	\item We propose the point-independent global attention module, which could learn global information from entire point clouds in an efficient way.
	\item We propose the point-dependent global attention module, which has a lower computational complexity than the existing non-local module~\cite{2018nlnet}. Besides, we propose a novel point-adaptive aggregation block.
	\item We propose the GA-Net consisting of the point-independent global attention module and the point-dependent global attention module. Extensive experimental results on point cloud semantic segmentation demonstrate that the proposed model surpass state-of-the-art methods in most cases.
\end{itemize}

\section{Global Attention Network}

%In this section, we firstly introduce the architecture of proposed GA-Net. Then, we describe the details of the point-independent global attention module and the point-dependent global attention module respectively.

\subsection{Architecture}
%We illustrate our end-to-end GA-Net framework in Figure \ref{fig1}.
As shown in Figure \ref{fig1}, our end-to-end GA-Net consists of three modules, including a feature extractor, a point-independent global attention module, and a point-dependent global attention module. The feature extractor is a U-Net based on RandLA-Net~\cite{2020randlanet}. 
RandLA-Net takes the entire point clouds as input and is able to efficiently infer per-point semantics in a single pass. 
Besides, the performances of RandLA-Net is somewhat state-of-the-art on several benchmarks.

When a 3D point cloud $\mathbf{P} = \{\boldsymbol{p}_1, \boldsymbol{p}_2, ..., \boldsymbol{p}_N\}  \in \mathbb{R}^{N \times (3 + d)}$ is given, where $N$ is the number of points and $3 + d$ denotes the xyz-dimension and additional properties (e.g., $d = 3$ for RGB or normal information), we firstly send $\mathbf{P}$ to the feature extractor to construct its high-level representation $\mathbf{F} = \{\boldsymbol{f}_1, \boldsymbol{f}_2, ..., \boldsymbol{f}_N\} \in \mathbb{R}^{N \times C}$ where $C$ is the dimension of high-level features. Secondly, we send $\mathbf{F}$ to the point-independent global attention module to get feature map $\mathbf{G} = \{\boldsymbol{g}_1, \boldsymbol{g}_2, ..., \boldsymbol{g}_N\}  \in \mathbb{R}^{N \times C}$. Then, we feed $\mathbf{G}$ into the point-dependent global attention module to generate the context-aware feature map $\mathbf{X} = \{\boldsymbol{x}_1, \boldsymbol{x}_2, ..., \boldsymbol{x}_N\}  \in \mathbb{R}^{N \times C}$. Lastly, the feature map $\mathbf{X}$ is fed into fully-connected(FC) layers for label assignment.

%We illustrate our end-to-end GA-Net framework in Figure \ref{fig1}. Given a 3D point cloud $\mathbf{P} = \{\boldsymbol{p}_1, \boldsymbol{p}_2, ..., \boldsymbol{p}_N\}  \in \mathbb{R}^{N \times (3 + d)}$, where $N$ is the number of points and $3 + d$ denotes the xyz-dimension and additional properties (e.g., $d = 3$ for RGB or normal information), through feature extraction, we construct their high-level representation, forming high-dimensional features $\mathbf{F} = \{\boldsymbol{f}_1, \boldsymbol{f}_2, ..., \boldsymbol{f}_N\}  \in \mathbb{R}^{N \times C}$ where $C$ is the dimension of high-level features. The feature extractor is a U-Net based on RandLA-Net~\cite{2020randlanet}. RandLA-Net takes the entire point clouds as input and is able to efficiently infer per-point semantics in a single pass. Besides, the performances of RandLA-Net is somewhat state-of-the-art on several benchmarks. After feature extraction, we introduce two modules named the point-independent global attention module and the point-dependent global attention module, which generate context-aware feature maps $\mathbf{G} = \{\boldsymbol{g}_1, \boldsymbol{g}_2, ..., \boldsymbol{g}_N\}  \in \mathbb{R}^{N \times C}$ and $\mathbf{X} = \{\boldsymbol{x}_1, \boldsymbol{x}_2, ..., \boldsymbol{x}_N\}  \in \mathbb{R}^{N \times C}$ respectively. Finally, the feature map $\mathbf{X}$ is fed into fully-connected(FC) layers for label assignment.

\subsection{Point-independent Global Attention Module}
For aggregating point-independent global information efficiently, we introduce the point-independent global attention module. Firstly, we reduce the number of feature channels to one by sending $\mathbf{F}$ to a shared MLP. Then we perform a normalization using softmax to obtain the global attention map $\boldsymbol{w} \in \mathbb{R}^{N}$:
\begin{equation}
	\begin{split}
		&w_i = \frac{\textup{exp}(\textup{MLP}(\boldsymbol{f}_i))}{\sum_{j=1}^N \textup{exp}(\textup{MLP}(\boldsymbol{f}_j))} \\
	\end{split}
	\label{equ1}
\end{equation}
where $w_{i}$ is the $i$-th element of $\boldsymbol{w}$. After getting the attention map, we firstly perform attention pooling for all points to obtain the global contextual feature. Then two FC layers are applied to learn channel-wise dependencies. For reducing the difficulty of optimization, we add a layer normalization(LN) between the two FC layers (before ReLU). Finally, the output feature is stacked into N copies and connected with $\mathbf{F}$ by the element-wise multiplication to get the feature map $\mathbf{G}$. The whole procedure of the point-independent global attention is formulated as:
\begin{equation}
	\begin{split}
		&\boldsymbol{g}_i = \boldsymbol{f}_i \odot \textup{FC}(\textup{ReLU}(\textup{LN}(\textup{FC}(\sum_{j=1}^N w_i \boldsymbol{f}_i)))) \\
	\end{split}
	\label{equ2}
\end{equation}
where `$\odot$' represents the element-wise multiplication. 

Overall, this module can extract not only the point-independent global information efficiently, but also the relationships between different feature channels.

\begin{figure}[t]
	\centering
	\includegraphics[width=0.9 \columnwidth]{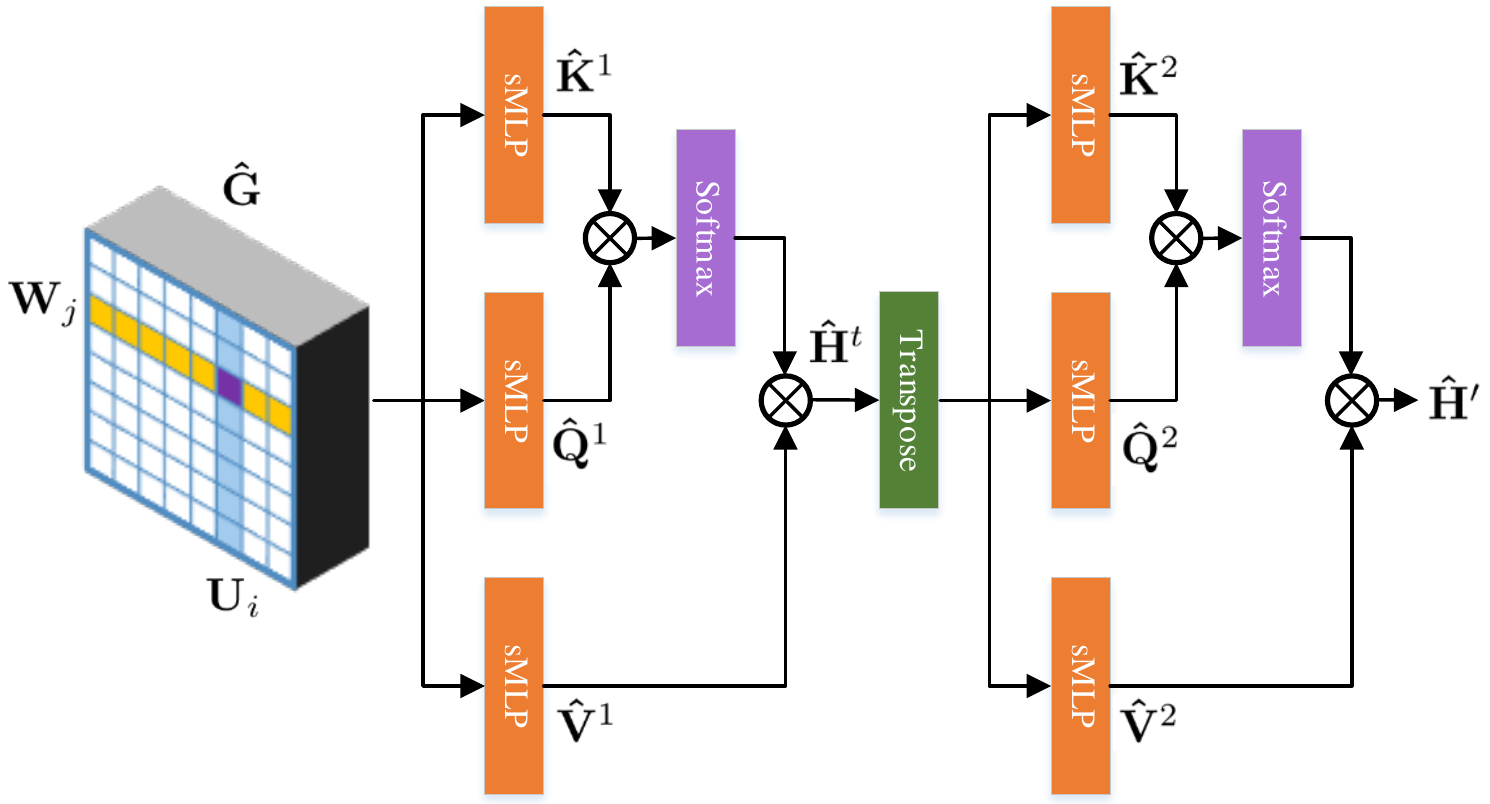}
	\caption{Architecture of the random cross attention block.}
	\label{fig2}
\end{figure}

\begin{table*}[t]
	\begin{center}
		\caption{Semantic segmentation results (\%) on the Semantic3D dataset (semantic-8).} \label{tab_semantic3d}
		\resizebox{2\columnwidth}{!}{
			\begin{tabular}{|c|c|c|c|c|c|c|c|c|c|c|}
				\hline
				\textbf{Method}         &\textbf{mIoU}    &\textbf{OA}    &man-made terrain    &natural terrain    &high vegetation    &low vegetation    &buildings    &hard scape    &scanning artefacts    &cars\\
				\hline
				PointNet++~\cite{2017pointnet++}     &63.1	&85.7	&81.9	&78.1	&64.3	&51.7	&75.9	&36.4	&43.7	&72.6\\
				EC-PointNet~\cite{2019ecpointnet}    &64.4	&89.6	&91.1	&69.5	&65.0	&56.0	&89.7	&30.0	&43.8	&69.7\\
				SnapNet~\cite{2017snapnet}           &67.4	&91.0	&89.6	&79.5	&\textbf{74.8}	&56.1	&90.9	&36.5	&34.3	&77.2\\
				PointGCR~\cite{2020pointgcr}         &69.5	&92.1	&93.8	&80.0	&64.4	&66.4	&93.2	&39.2	&34.3	&85.3\\
				PointCE~\cite{2020sce}               &71.0	&92.3	&92.4	&79.6	&72.7	&62.0	&93.7	&40.6	&44.6	&82.5\\
				RandLA-Net~\cite{2020randlanet}	     &71.9	&94.1	&95.9	&88.3	&65.5	&61.7	&95.9	&\textbf{50.0}	&27.6	&90.2\\
				Fast-PCR~\cite{2019fastpcr}          &72.0	&90.6	&86.4	&70.3	&69.5	&\textbf{68.0}	&\textbf{96.9}	&43.4	&\textbf{52.3}	&89.5\\
				%SPGraph~\cite{2018spgraph}           &\textbf{76.2}	&92.9	&91.5	&75.6	&\textbf{78.3}	&\textbf{71.7}	&94.4	&\textbf{56.8}	&\textbf{52.9}	&88.4\\
				%FKAConv~\cite{2020fkaconv}           &94.1	&\textbf{74.6}	&94.7	&85.2	&77.4	&70.4	&94.0	&52.9	&29.4	&\textbf{92.6}\\
				GA-Net(ours)                        &\textbf{74.3}	&\textbf{94.6}	&\textbf{96.7}	&\textbf{91.5}	&63.3	&61.7	&96.1	&45.0	&49.1	&\textbf{91.3}\\
				\hline
			\end{tabular}
		}
	\end{center}
\end{table*}

\begin{table*}[t]
	\begin{center}
		\caption{Semantic segmentation results (\%) on the S3DIS dataset (Area-5).} \label{tab_s3dis_1}
		\resizebox{2\columnwidth}{!}{
			\begin{tabular}{|c|c|c|c|c|c|c|c|c|c|c|c|c|c|c|c|}
				\hline
				\textbf{Method}	&\textbf{mIoU}	&\textbf{OA}	&ceiling	&floor	&wall	&beam	&column	&window	&door	&chair	&table	&bookcase	&sofa	&board	&clutter\\
				\hline
				%PointNet~\cite{2017pointnet}    	   &41.1 	&-	&88.8 	&97.3 	&69.8 	&0.1 	&3.9 	&46.3 	&10.8 	&52.6 	&58.9 	&40.3 	&5.9 	&26.4 	&33.2\\ 
				PointNet++~\cite{2017pointnet++}	   &50.0 	&-	&90.8 	&96.5 	&74.1 	&0.0 	&5.8 	&43.6 	&25.4 	&69.2 	&76.9 	&21.5 	&55.6 	&49.3 	&41.9\\ 
				PointGCR~\cite{2020pointgcr}	       &54.4 	&-	&90.7 	&96.1 	&74.9 	&\textbf{0.1} 	&16.1 	&50.2 	&32.3 	&69.0 	&78.1 	&41.3 	&60.7 	&53.8 	&43.8\\ 
				%SPGraph~\cite{2018spgraph}	           &58.0 	&86.4	&89.4 	&96.9 	&78.1 	&0.0 	&\textbf{42.8} 	&48.9 	&61.6 	&\textbf{84.7} 	&75.4 	&69.8 	&52.6 	&2.1 	&52.2\\ 
				PointCNN~\cite{2018pointcnn}	       &57.3 	&86.0	&92.3 	&98.2 	&79.4 	&0.0 	&17.6 	&22.8 	&\textbf{62.1} 	&74.4 	&80.6 	&31.7 	&66.7 	&62.1 	&\textbf{56.7}\\ 
				PointWeb~\cite{2019pointweb}	       &60.3 	&87.0	&92.0 	&\textbf{98.5} 	&79.4 	&0.0 	&21.1 	&59.7 	&34.8 	&76.3 	&\textbf{88.3} 	&46.9 	&69.3 	&64.9 	&52.5\\ 
				RandLA-Net~\cite{2020randlanet}	       &61.6 	&86.7	&91.2 	&95.6 	&79.5 	&0.0 	&20.6 	&59.9 	&43.4 	&76.5 	&82.8 	&60.8 	&70.4 	&67.9 	&52.0\\ 
				GACNet~\cite{2019gacnet}	           &62.9 	&87.8	&92.3 	&98.3 	&81.9 	&0.0 	&20.4 	&59.1 	&40.9 	&\textbf{85.8} 	&78.5 	&70.8 	&61.7 	&\textbf{74.7} 	&52.8\\ 
				Point2Node~\cite{2020point2node}	   &63.0 	&\textbf{88.8}	&\textbf{93.9} 	&98.3 	&\textbf{83.3} 	&0.0 	&\textbf{35.7} 	&55.3 	&58.8 	&79.5 	&84.7 	&44.1 	&\textbf{71.1} 	&58.7 	&55.2\\ 
				GA-Net(ours)	                       &\textbf{63.7} 	&87.6	&92.9 	&97.8 	&81.3 	&0.0 	&27.8 	&\textbf{60.3} 	&41.7 	&78.3 	&86.7 	&\textbf{71.4} 	&69.9 	&65.8 	&53.9\\ 
				\hline
			\end{tabular}
		}
	\end{center}
\end{table*}

\subsection{Point-dependent Global Attention Module}
To model long-range point-dependent dependencies with a lower computational costs and memory, we introduce a point-dependent global attention module. As shown in Figure \ref{fig1}, we firstly reorder 3D points by random sampling. Then, the feature map $\mathbf{G}$ is sent to two random cross attention blocks to produce $\mathbf{H}' \in \mathbb{R}^{N \times C}$ and $\mathbf{H}'' \in \mathbb{R}^{N \times C}$ respectively. Lastly, the features $\mathbf{H}''$ and $\mathbf{G}$ are sent to a point-adaptive aggregation block for aggregating more discriminate feature map $\mathbf{X}$.

\textbf{Reordering point clouds.} The whole point cloud can be divided into many subsets, where each subset is obtained through uniformly random sampling. Specifically, if the number of points in each subset is set to $k_1$, the number of a point cloud with $N$ will firstly be expanded to $\hat{N}$, making $\hat{N} = k_1 \times  k_2$ ($k_2 \in \mathbb{Z}$), to get the expanded point cloud $\mathbf{\hat{P}}$ where the expanded points are randomly sampled from the original point cloud. Then we sample $k_2$ times from the point cloud $\mathbf{\hat{P}}$, each of which contains $k_1$ randomly sampled points from the remaining unsampled points. So the matrix $\mathbf{\hat{P}}$ can be reshaped as $ \mathbf{\hat{P}} \in \mathbb{R}^{k_2 \times k_1 \times (3 + d)}$. If transposing the first dimension and the second dimension of this matrix, it can also be explained that the point cloud consists of $k_1$ subsets with $k_2$ random-sampled points. As seen in Figure \ref{fig2}, each point is contained in a subset $\mathbf{U}_i(i = 1, 2, ..., k_2)$ of $k_1$ points and a subset $\mathbf{W}_j(j = 1, 2, ..., k_1)$ of $k_2$ points. 

\textbf{Random cross attention block.} After reordering the point cloud, we design a novel random cross attention block to achieve non-local attention in a more efficient way. This block consists of a two-pass procedure which can be seen in Figure \ref{fig2}. According to $\mathbf{\hat{P}}$, the feature map $\mathbf{G}$ can be expanded to $\mathbf{\hat{G}}$, and then reshaped to $\mathbf{\hat{G}} \in \mathbb{R}^{k_2 \times k_1 \times C}$. 

In the first step, we feed $\mathbf{\hat{G}}$ into three shared MLPs to obtain the key feature map $\mathbf{\hat{K}}^1 = \{\mathbf{\hat{K}}_1^1, \mathbf{\hat{K}}_2^1, ..., \mathbf{\hat{K}}_{k_2}^1\} \in \mathbb{R}^{k_2 \times k_1 \times C}$, the query feature map $\mathbf{\hat{Q}}^1 = \{\mathbf{\hat{Q}}_1^1, \mathbf{\hat{Q}}_2^1, ..., \mathbf{\hat{Q}}_{k_2}^1\} \in \mathbb{R}^{k_2 \times k_1 \times C}$ and the value feature map $\mathbf{\hat{V}}^1 = \{\mathbf{\hat{V}}_1^1, \mathbf{\hat{V}}_2^1, ..., \mathbf{\hat{V}}_{k_2}^1\} \in \mathbb{R}^{k_2 \times k_1 \times C}$, respectively. The output feature map of the first step is $\mathbf{\hat{H}}^t = \{\mathbf{\hat{H}}_1^t, \mathbf{\hat{H}}_2^t, ..., \mathbf{\hat{H}}_{k_2}^t\} \in \mathbb{R}^{k_2 \times k_1 \times C}$. Specifically, for each subset $\mathbf{U}_i$, we use multiplication between $\mathbf{\hat{K}}_i^1$ and the transpose of $\mathbf{\hat{Q}}_i^1$ with a softmax to produce a self-attention map. Then $\mathbf{\hat{H}}_i^t$ is produced by multiplying between the self-attention map and $\mathbf{\hat{V}}_i^1$. The first step can be formulated as:
\begin{equation}
	\begin{split}
		&\mathbf{\hat{H}}_i^t = \textup{softmax}(\mathbf{\hat{K}}_i^1\mathbf{\hat{Q}}_i^{1 \top})\mathbf{\hat{V}}_i^1 \\
	\end{split}
	\label{equ3}
\end{equation}

In the second step, the feature map $\mathbf{\hat{H}}^t$ is firstly transposed on the first dimension and the second dimension, then similarly sent to three shared MLPs to obtain the key feature map $\mathbf{\hat{K}}^2 = \{\mathbf{\hat{K}}_1^2, \mathbf{\hat{K}}_2^2, ..., \mathbf{\hat{K}}_{k_1}^2\} \in \mathbb{R}^{k_1 \times k_2 \times C}$, the query feature map $\mathbf{\hat{Q}}^2 = \{\mathbf{\hat{Q}}_1^2, \mathbf{\hat{Q}}_2^2, ..., \mathbf{\hat{Q}}_{k_1}^2\} \in \mathbb{R}^{k_1 \times k_2 \times C}$ and the value feature map $\mathbf{\hat{V}}^2 = \{\mathbf{\hat{V}}_1^2, \mathbf{\hat{V}}_2^2, ..., \mathbf{\hat{V}}_{k_1}^2\} \in \mathbb{R}^{k_1 \times k_2 \times C}$, respectively. The output feature map of the second step is $\mathbf{\hat{H}}' = \{\mathbf{\hat{H}}'_1, \mathbf{\hat{H}}'_2, ..., \mathbf{\hat{H}}'_{k_1}\} \in \mathbb{R}^{k_1 \times k_2 \times C}$. For each subset $\mathbf{W}_j$, we apply multiplication between $\mathbf{\hat{K}}_j^2$ and the transpose of $\mathbf{\hat{Q}}_j^2$ with a softmax to produce a self-attention map. Then $\mathbf{\hat{H}}'_j$ is calculated by multiplying between the self-attention map and $\mathbf{\hat{V}}_j^2$. The second step can be formulated as:
\begin{equation}
	\begin{split}
		&\mathbf{\hat{H}}'_j = \textup{softmax}(\mathbf{\hat{K}}_j^2\mathbf{\hat{Q}}_j^{2 \top})\mathbf{\hat{V}}_j^2 \\
	\end{split}
	\label{equ4}
\end{equation}

After obtaining $\mathbf{\hat{H}}'$, we remove the expanded points and reshape it to $\mathbf{H}' \in \mathbb{R}^{N \times C}$. Then the random cross attention block is repeated again to get the enhanced context-aware feature map $\mathbf{H}''$. Theoretically, the attention maps of each point predicted by a random cross attention block only have about $2\sqrt{N}$ weights (if $k_1 = k_2 \approx \sqrt{N}$) which are much less then $N$ in the non-local module~\cite{2018nlnet}, leading to reduce the computational complexity from $\mathcal{O}(N^2C)$ to $\mathcal{O}(N\sqrt{N}C)$.

\textbf{Point-adaptive aggregation block.} After two random cross attention blocks, how to connect the context-aware feature map $\mathbf{H}''$ and input features $\mathbf{G}$ to get $\mathbf{X}$ needs to be solved. Traditional DNNs aggregate different features using linear aggregations such as skip connection. But linear aggregations are not data-adaptive. To better reflect the characteristics of different points and make the aggregation data-aware, we propose a simple point-adaptive aggregation block. We use shared MLPs with one output channel for the two feature maps to produce two weight maps. Then a weighted summation to the two feature maps is performed to obtain $\mathbf{X}$. The whole procedure can be formulated as:
\begin{equation}
	\begin{split}
		\boldsymbol{x}_i = &\frac{\textup{exp}(\textup{MLP}(\boldsymbol{h}''_i))}{\textup{exp}(\textup{MLP}(\boldsymbol{h}''_i)) + \textup{exp}(\textup{MLP}(\boldsymbol{g}_i))} \boldsymbol{h}''_i + \\
		&\frac{\textup{exp}(\textup{MLP}(\boldsymbol{g}_i))}{\textup{exp}(\textup{MLP}(\boldsymbol{h}''_i)) + \textup{exp}(\textup{MLP}(\boldsymbol{g}_i))} \boldsymbol{g}_i \\
	\end{split}
	\label{equ5}
\end{equation}
where $\boldsymbol{h}''_i$ is the $i$-th component of $\mathbf{H}''$.

\section{Experimental Results}

%In this section, we firstly introduce the details of datasets and implementation. Secondly, we compare the performances between the proposed GA-Net and several state-of-the-art methods on three public datasets. Lastly, we provide an ablation analysis.

\subsection{Datasets and Implementation Details}
The proposed GA-Net is evaluated on three datasets, including Semantic3D~\cite{2017semantic3d}, S3DIS~\cite{2016s3dis}, and ScanNet~\cite{2017scannet}. 
%The Semantic3D is a outdoor dataset and the other two are indoor datasets.
%In addition to the coordinates of 3D points, we only consider the color information as inputs. 
In the feature extractor, the U-Net parameters are consistent with the model before the FC layers in RandLA-Net~\cite{2020randlanet}. 
The output dimensionalities of all the layers in the proposed two modules are 16. 
The FC has two layers, where the output dimensionalities are 64 and 32 respectively. 
We train our GA-Net using the Adam optimizer with initial learning rate 0.01 and batchsize 6 for 100 epochs.

\subsection{Results on the Semantic3D Dataset}
The Semantic3d dataset contains 30 outdoor scenes, of which 15 are used as training and the remaining are used as online testing. Each point cloud has up to $10^8$ points with RGB and intensity values, and is labeled from 8 semantic categories. We conducted experiments on the semantic-8 challenge. To make a fair comparison, we calculated the mean of class-wise intersection over union (mIoU) and the overall point-wise accuracy (OA) following Fast-PCR~\cite{2019fastpcr}.

We compared our GA-Net to several state-of-the-art methods according to the Semantic3D online evaluation website, as summarized in Table \ref{tab_semantic3d}. Here, we take RandLA-Net~\cite{2020randlanet} as our baseline. In Table \ref{tab_semantic3d}, our GA-Net outperforms its baseline by $2.4\%$ in terms of mIoU and $0.5\%$ in terms of OA. The comparative results also show us that our method achieves best on both metrics, due to its more effective and efficient global contextual-feature learning. 
%In particular, the mIoU of the man-made terrain and natural terrain categories outperforms all the current methods, because our method can better distinguish categories that are similar in local geometry but distinct in global distribution.
Our method outperforms the comparative methods for segmenting the objects (e.g. man-made terrain and natural terrain) which are of a relatively bigger size, but achieves lower performances than several comparative methods (particularly Fast-PCR~\cite{2019fastpcr}) for segmenting the objects (e.g. high vegetation and the low vegetation) which are of a relatively smaller size. This is mainly because: global features could generally reflect the characteristics of large-sized objects, while it generally needs local features for discriminating small-sized objects. 

\subsection{Results on the S3DIS Dataset}
The S3DIS dataset contains 6 areas with 271 rooms in buildings. Each point, with xyz coordinates and RGB features, is annotated with one semantic label from 13 categories. We conducted our experiments in Area-5 validation. 
%Since the fifth area does not overlap with other areas, experiments on Area-5 could better reflect the generalization ability of the framework. 
To make a fair comparison, the evaluation metrics we used are mIoU and OA following GACNet~\cite{2019gacnet}.
The quantitative results are reported in Table \ref{tab_s3dis_1}. The proposed GA-Net has a highest mIoU among all the competitive methods on Area-5, which demonstrates that our method has a good ability to learn the global information. Comparing with the baseline, our method improves mIoU by $2.1\%$ in terms of mIoU and $0.9\%$ in terms of OA.
%Our method performs particularly well in the categories of windows and bookcases, indicating that our method can better explore the geometric characteristics of these two object classes. 

\begin{table}[t]
	\begin{center}
		\caption{Semantic segmentation results (\%) on the ScanNet dataset.} \label{tab_scannet}
		\resizebox{0.55\columnwidth}{!}{
			\begin{tabular}{|c|c|}
				\hline
				\textbf{Method}	        &\textbf{Per-voxel accuracy}\\
				\hline
				%PointNet~\cite{2017pointnet}	        &73.9\\
				PointNet++~\cite{2017pointnet++}	    &84.5\\
				PointCNN~\cite{2018pointcnn}	        &85.1\\
				PointGCR~\cite{2020pointgcr}	        &85.3\\
				%ACNN~\cite{2019acnn}	                &85.4\\
				PointWeb~\cite{2019pointweb}	        &85.9\\
				RandLA-Net~\cite{2020randlanet}	        &86.1\\
				PointSIFT~\cite{2018pointsift}	        &86.2\\
				Point2Node~\cite{2020point2node}	    &86.3\\
				GA-Net(ours)	                        &\textbf{86.6}\\
				\hline
			\end{tabular}
		}
	\end{center}
\end{table}

\begin{table}[t]
	\begin{center}
		\caption{Ablation study of the sub-modules on the S3DIS dataset (Area-5).} \label{tab_ablation}
		\resizebox{1\columnwidth}{!}{
			\begin{tabular}{|c|c|c|c|c|c|}
				\hline
				\textbf{Method}	 &\textbf{mIoU}(\%)	&FLOPS	&params	&memory(GB) &time(ms)\\
				\hline
				baseline	      &61.6	&31333891	&4993141	&15.4	&121\\
				1-RCAB+plus	      &62.6	&31378063	&4999861	&16.8	&142\\
				1-RCAB+PAB	      &62.9	&31378527	&4999931	&17.1	&146\\
				2-RCAB+PAB	      &63.0	&31422699	&5006651	&18.7	&157\\
				GA-Net(ours)	  &\textbf{63.7}	&31434393	&5009918	&18.8	&160\\
				\hline
			\end{tabular}
		}
	\end{center}
\end{table}

\subsection{Results on the ScanNet Dataset}
The ScanNet dataset contains 1,513 scanned and reconstructed indoor scenes, which provides a 1,201/312 scene split for training and testing. 20 categories are provided for evaluation. To make a fair comparison, we reported the per-voxel accuracy following Point2Node~\cite{2020point2node}. Table \ref{tab_scannet} shows the comparisons between our GA-Net and other competitive methods. Our method achieves the state-of-the-art performance, which improves on baseline by $0.5\%$, due to its more effective and efficient feature learning of long-range dependencies.

\subsection{Ablation Study}
For ablation study, we stacked the proposed sub-modules on the baseline step-to-step to prove the effectiveness of our method. Specifically, the comparing experiments are (1) baseline, (2) adding one random cross attention block (RCAB) and aggregating features with a plus operation, denoted as ``1-RCAB+plus", (3) adding one random cross attention block and aggregating features by point-adaptive aggregation block (PAB), denoted as ``1-RCAB+PAB", (4) adding two random cross attention blocks and the rest is consistent with (3), denoted as ``2-RCAB+PAB", and (5) our proposed GA-Net. 
Our baseline method employs a U-Net based on RandLA-Net~\cite{2020randlanet}. 
We conducted ablation study on Area-5 of the S3DIS with the evaluation metric mIoU. Besides, we also made statistics on the floating-point operations per second (FLOPs), number of parameters, computing memory, and computing time to prove the efficiency of our method.

As shown in Table \ref{tab_ablation}, ``1-RCAB+plus" performing better than baseline demonstrates that the importance of exploring point-dependent global information. More interestingly, the result of ``1-RCAB+PAB" achieves better than ``1-RCAB+plus" , which may be attributed to the data-aware aggregating method. ``2-RCAB+PAB" performing better than ``1-RCAB+PAB" indicates that stacking more random cross attention blocks benefits context-aware feature learning. Our proposed method achieves best, indicating that combining two kinds of global information can further improve results. Furthermore, the last four columns in Table \ref{tab_ablation} demonstrate that our method does not require much calculation, memory and time. But stacking more random cross attention blocks results in waste of resources, so only two of it is considered.

\textbf{Remark.} It is pointed out that the computational complexity of the non-local module~\cite{2018nlnet} is much greater than the random cross attention block as mentioned in Section \uppercase\expandafter{\romannumeral2}-C. Hence we did not train the network consisting of the baseline and the non-local module since our GPU memory(32GB) can not meet the storage requirements.

\section{Conclusion}

For obtaining long-range contextual information of 3D point clouds, we propose a global attention network, called GA-Net, which consists of a point-independent global attention module, and a point-dependent global attention module. These two modules can obtain global information in an efficient way. 
%In addition, a point-adaptive aggregation block is proposed to aggregate more discriminate features. 
Extensive experiments on three point cloud benchmarks demonstrate that our method outperforms state-of-the-art methods in most cases.

\balance
%\section*{References}


\begin{thebibliography}{}
\providecommand{\url}[1]{#1}
\csname url@samestyle\endcsname
\providecommand{\newblock}{\relax}
\providecommand{\bibinfo}[2]{#2}
\providecommand{\BIBentrySTDinterwordspacing}{\spaceskip=0pt\relax}
\providecommand{\BIBentryALTinterwordstretchfactor}{4}
\providecommand{\BIBentryALTinterwordspacing}{\spaceskip=\fontdimen2\font plus
\BIBentryALTinterwordstretchfactor\fontdimen3\font minus
  \fontdimen4\font\relax}
\providecommand{\BIBforeignlanguage}[2]{{%
\expandafter\ifx\csname l@#1\endcsname\relax
\typeout{** WARNING: IEEEtran.bst: No hyphenation pattern has been}%
\typeout{** loaded for the language `#1'. Using the pattern for}%
\typeout{** the default language instead.}%
\else
\language=\csname l@#1\endcsname
\fi
#2}}
\providecommand{\BIBdecl}{\relax}
\BIBdecl

\end{thebibliography}


\begin{thebibliography}{34}
\bibitem{2016s3dis}
I.~Armeni, O.~Sener, A.~Zamir, H.~Jiang, I.~Brilakis, M.~Fischer, and
S.~Savarese, ``3d semantic parsing of large-scale indoor spaces,'' in
\emph{Proceedings of the IEEE Conference on Computer Vision and Pattern Recognition(CVPR)}, 2016, pp. 1534--1543.

\bibitem{2017scannet}
A.~Dai, A.~X. Chang, M.~Savva, M.~Halber, T.~Funkhouser, and M.~Nie{\ss}ner,
``Scannet: Richly-annotated 3d reconstructions of indoor scenes,'' in
\emph{Proceedings of the IEEE Conference on Computer Vision and Pattern Recognition(CVPR)}, 2017, pp. 2432--2443.

\bibitem{2017semantic3d}
T.~Hackel, N.~Savinov, L.~Ladicky, J.~D. Wegner, K.~Schindler, and
M.~Pollefeys, ``Semantic3d.net: A new large-scale point cloud classification
benchmark,'' \emph{arXiv:1704.03847}, 2017.	
	
\bibitem{2017pointnet}
C.~R. Qi, H.~Su, K.~Mo, and L.~J. Guibas, ``Pointnet: Deep learning on point
sets for 3d classification and segmentation,'' in \emph{Proceedings of the IEEE Conference on Computer Vision and Pattern Recognition(CVPR)}, 2017, pp.
652--660.

\bibitem{2017pointnet++}
C.~R. Qi, L.~Yi, H.~Su, and L.~J. Guibas, ``Pointnet++: Deep hierarchical
feature learning on point sets in a metric space,'' in \emph{Proceedings of the International Conference on Neural Information Processing Systems(NeurIPS)}, 2017,
pp. 5099--5108.

\bibitem{2017snapnet}
A.~Boulch, B.~L. Saux, and N.~Audebert, ``Unstructured point cloud semantic
labeling using deep segmentation networks,'' in \emph{Proceedings of the Workshop on 3D Object Retrieval(3DORW)}, 2017.

\bibitem{2018dgcnn}
Y.~Wang, Y.~Sun, Z.~Liu, S.~E. Sarma, M.~M. Bronstein, and J.~M. Solomon,
``Dynamic graph cnn for learning on point clouds,'' \emph{Acm Transactions On
	Graphics (TOG)}, 2019, pp. 1--12.

\bibitem{2018pointcnn}
Y.~Li, R.~Bu, M.~Sun, W.~Wu, X.~Di, and B.~Chen, ``Pointcnn: Convolution on
x-transformed points,'' in \emph{Proceedings of the International Conference on Neural Information Processing Systems(NeurIPS)}, 2018, pp. 820--830.

\bibitem{2018pointsift}
M.~Jiang, Y.~Wu, and C.~Lu, ``Pointsift: A sift-like network module for 3d
point cloud semantic segmentation,'' \emph{arXiv: 1807.00652},
2018.

\bibitem{2018spidercnn}
Y.~Xu, T.~Fan, M.~Xu, L.~Zeng, and Y.~Qiao, ``Spidercnn: Deep learning on point
sets with parameterized convolutional filters,'' in \emph{Proceedings of the European Conference on Computer Vision (ECCV)}, 2018, pp. 87--102.

\bibitem{2018nlnet}
X.~Wang, R.~Girshick, A.~Gupta, and K.~He, ``Non-local neural networks,'' in
\emph{Proceedings of the IEEE Conference on Computer Vision and Pattern Recognition(CVPR)}, 2018, pp. 7794--7803.

\bibitem{2019gacnet}
L.~Wang, Y.~Huang, Y.~Hou, S.~Zhang, and J.~Shan, ``Graph attention convolution
for point cloud semantic segmentation,'' in \emph{Proceedings of the IEEE Conference on Computer Vision and Pattern Recognition(CVPR)}, 2019, pp.
10296--10305.

\bibitem{2019pointweb}
H.~Zhao, L.~Jiang, C.-W. Fu, and J.~Jia, ``Pointweb: Enhancing local
neighborhood features for point cloud processing,'' in \emph{Proceedings of the IEEE Conference on Computer Vision and Pattern Recognition(CVPR)}, 2019, pp.
5565--5573.

\bibitem{2019ccnet}
Z.~Huang, X.~Wang, L.~Huang, C.~Huang, Y.~Wei, H.~Shi, and W.~Liu, ``Ccnet:
Criss-cross attention for semantic segmentation,'' in \emph{Proceedings of the IEEE/CVF International Conference on Computer Vision(ICCV)}, 2019, pp.
603--612.

\bibitem{2019gcnet}
Y.~Cao, J.~Xu, S.~Lin, F.~Wei, and H.~Hu, ``Gcnet: Non-local networks meet
squeeze-excitation networks and beyond,'' in \emph{Proceedings of the IEEE/CVF International Conference on Computer Vision Workshops(ICCVW)}, 2019, pp.
1971--1980.

\bibitem{2019fastpcr}
G.~Truong, S.~Z. Gilani, S.~M.~S. Islam, and D.~Suter, ``Fast point cloud
registration using semantic segmentation,'' \emph{Digital Image
	Computing: Techniques and Applications (DICTA)}, 2019, pp. 1--8.

\bibitem{2019ecpointnet}
J.~Contreras and J.~Denzler, ``Edge-convolution point net for semantic
segmentation of large-scale point clouds,'' \emph{IEEE
	International Geoscience and Remote Sensing Symposium(IGARSS)}, 2019, pp. 5236--5239.

\bibitem{2020sce}
H.~Liu, Y.~Guo, Y.~Ma, Y.~Lei, and G.~Wen, ``Semantic context encoding for
accurate 3d point cloud segmentation,'' \emph{IEEE Transactions on
	Multimedia(TMM)}, 2020.

\bibitem{2020randlanet}
Q.~Hu, B.~Yang, L.~Xie, S.~Rosa, Y.~Guo, Z.~Wang, A.~Trigoni, and A.~Markham,
``Randla-net: Efficient semantic segmentation of large-scale point clouds,''
in \emph{Proceedings of the IEEE Conference on Computer Vision and Pattern Recognition(CVPR)}, 2020, pp. 11105--11114.

\bibitem{2020point2node}
W.~Han, C.~Wen, C.~Wang, X.~Li, and Q.~Li, ``Point2node: Correlation learning
of dynamic-node for point cloud feature modeling,'' in \emph{Proceedings of the AAAI Conference on Artificial Intelligence(AAAI)}, 2020, pp. 10925--10932.

\bibitem{2020mprm}
J.~Wei, G.~Lin, K.-H. Yap, T.-Y. Hung, and L.~Xie, ``Multi-path region mining
for weakly supervised 3d semantic segmentation on point clouds,'' in
\emph{Proceedings of the IEEE Conference on Computer Vision and Pattern Recognition(CVPR)}, 2020, pp. 4383--4392.

\bibitem{2020pointgcr}
Y.~Ma, Y.~Guo, H.~Liu, Y.~Lei, and G.-J. Wen, ``Global context reasoning for
semantic segmentation of 3d point clouds,'' in \emph{Proceedings of the IEEE/CVF Winter Conference on Applications of Computer Vision(WACV)}, 2020, pp.
2920--2929.

\bibitem{2020lrnnet}
W.~Jiang, Z.~Xie, Y.~Li, C.~Liu, and H.~Lu, ``Lrnnet: A light-weighted network
with efficient reduced non-local operation for real-time semantic
segmentation,'' in \emph{IEEE International Conference on Multimedia and Expo Workshops (ICMEW)}, 2020, pp. 1--6.

\bibitem{2021scfnet}
S.~Fan, Q.~Dong, F.~Zhu, Y.~Lv, P.~Ye, and F.~Wang, ``Scf-net: Learning spatial contextual features for large-scale point cloud segmentation,'' in \emph{Proceedings of the IEEE Conference on Computer Vision and Pattern Recognition(CVPR)}, 2021. 

\bibitem{2021rtn}
S.~Deng, B.~Liu, Q.~Dong, and Z.~Hu, ``Rotation transformation network: Learning view-invariant point cloud for classification and segmentation,'' in \emph{Proceedings of the IEEE International Conference on Multimedia and Expo (ICME)}, 2021. 


\end{thebibliography}
\end{document}